\documentclass[10pt,twocolumn,letterpaper]{article}

\usepackage[pagenumbers]{cvpr} 

\usepackage[utf8]{inputenc} 
\usepackage[T1]{fontenc}    

\definecolor{cvprblue}{rgb}{0.21,0.49,0.74}
\usepackage[pagebackref,breaklinks,colorlinks,allcolors=cvprblue]{hyperref}
\usepackage{wrapfig} 
\usepackage{graphicx}
\usepackage[utf8]{inputenc} 
\usepackage[T1]{fontenc}    
\usepackage{hyperref}       
\usepackage{url}            
\usepackage{booktabs}       
\usepackage{amsfonts}       
\usepackage{nicefrac}       
\usepackage{microtype}      
\usepackage{xcolor}         
\usepackage{hyperref}
\usepackage{url}
\usepackage[table]{xcolor}
\usepackage{booktabs}
\usepackage{multicol}
\usepackage[numbers]{natbib} 
\usepackage{amsmath}
\usepackage{pifont}
\usepackage{amssymb}
\usepackage{bbding}
\usepackage{pifont}
\usepackage{makecell}
\usepackage{multirow}
\usepackage{wrapfig}
\usepackage{enumitem}
\usepackage{adjustbox}

\newcolumntype{C}[1]{>{\centering\arraybackslash}m{#1}}
\definecolor{Gray}{gray}{0.93}
\definecolor{myblue}{RGB}{0, 152, 204}
\definecolor{myred}{RGB}{204, 10, 10}
\definecolor{citecolor}{HTML}{2980b9}
\definecolor{linkcolor}{HTML}{c0392b}
\definecolor{darkorange}{HTML}{FF8C00}
\definecolor{chocolate}{HTML}{D2691E}
\definecolor{darkgreen}{HTML}{006400}
\definecolor{darkblue}{HTML}{00008B}
\definecolor{mediumblue}{HTML}{0000CD}
\definecolor{dodgerblue}{HTML}{1E90FF}
\definecolor{royalblue}{HTML}{4169E1}
\definecolor{shadecolor}{RGB}{237,237,237}
\definecolor{backred}{RGB}{255, 190, 190}
\definecolor{backblue}{RGB}{210, 230, 250}
\usepackage{tcolorbox}
\definecolor{zrrgreen}{HTML}{008000}
\definecolor{zrrblue}{HTML}{4682B4}
\definecolor{zrrred}{HTML}{B22222}

\newcommand{\model}{\textit{d}VLM-AD}

\def\paperID{17467} 
\def\confName{CVPR}
\def\confYear{2026}

\title{\textit{d}VLM-AD: Enhance Diffusion Vision-Language-Model for Driving via Controllable Reasoning}

\author{
Yingzi Ma$^{1}$, 
Yulong Cao$^{2}$, 
Wenhao Ding$^{2}$, 
Shuibai Zhang$^{1}$, 
Yan Wang$^{2}$, \\
Boris Ivanovic$^{2}$,
Ming Jiang$^{1}$,
Marco Pavone$^{2,3}$,
Chaowei Xiao$^{2,4}$
\\[0.2cm]
$^1$University of Wisconsin-Madison~~~~$^2$NVIDIA~~~~$^3$Stanford University~~~~$^4$Johns Hopkins University \\
\href{https://dvlm-ad.github.io}{https://dvlm-ad.github.io}
}

\begin{document}
\maketitle
\begin{abstract}

The autonomous driving community is increasingly focused on addressing the challenges posed by out-of-distribution (OOD) driving scenarios. A dominant research trend seeks to enhance end-to-end (E2E) driving systems by integrating vision–language models (VLMs), leveraging their rich world knowledge and reasoning abilities to improve generalization across diverse environments. However, most existing VLMs or vision–language agents (VLAs) are built upon autoregressive (AR) models.
In this paper, we observe that existing AR-based VLMs--limited by causal attention and sequential token generation--often fail to maintain consistency and controllability between high-level reasoning and low-level planning. In contrast, recent discrete diffusion VLMs equipped with bidirectional attention exhibit superior controllability and reliability through iterative denoising. Building on these observations, we introduce \model{}, a diffusion-based vision–language model that unifies perception, structured reasoning, and low-level planning for end-to-end driving. 
Evaluated on nuScenes and WOD-E2E, \model{} yields more consistent reasoning–action pairs and achieves planning performance comparable to existing driving VLM/VLA systems despite a modest backbone, outperforming ARM-based baselines with a 9\% improvement in behavior–trajectory consistency and a 6\% increase in RFS on long-tail WOD-E2E scenarios. These results suggest a controllable and reliable pathway for scalable end-to-end driving.  \looseness=-1

\end{abstract}    
\section{Introduction}
\label{sec:intro}

\begin{figure}[!t]
    \centering
    \includegraphics[width=0.95\linewidth]{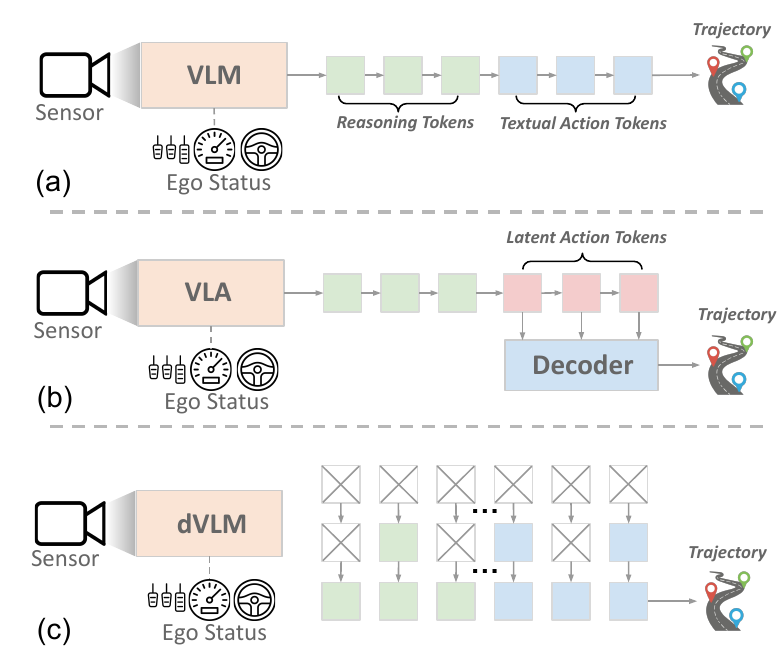}
    
    \caption{Comparison of end-to-end autonomous driving paradigms. (a) Autoregressive VLMs sequentially decode reasoning and textual action tokens, where each prediction depends on previous outputs, leading to accumulated exposure bias and limited global consistency. (b) VLAs introduce latent action tokens and a separate decoder to produce trajectories, but reasoning–action coupling remains implicit. (c) Our dVLM reformulate driving as an iterative denoising process that jointly refines reasoning and action representations under sensor and ego-state conditioning. This diffusion formulation eliminates left-to-right dependencies, enhances stability, and achieves stronger reasoning–action alignment within the end-to-end autonomous driving system.}
    \label{fig:comparison}
    \vspace{-5mm}
\end{figure}

\begin{figure*}[!t]
    \centering
    \includegraphics[width=0.95\linewidth,
    trim=0 0pt 0 0pt, 
    clip]{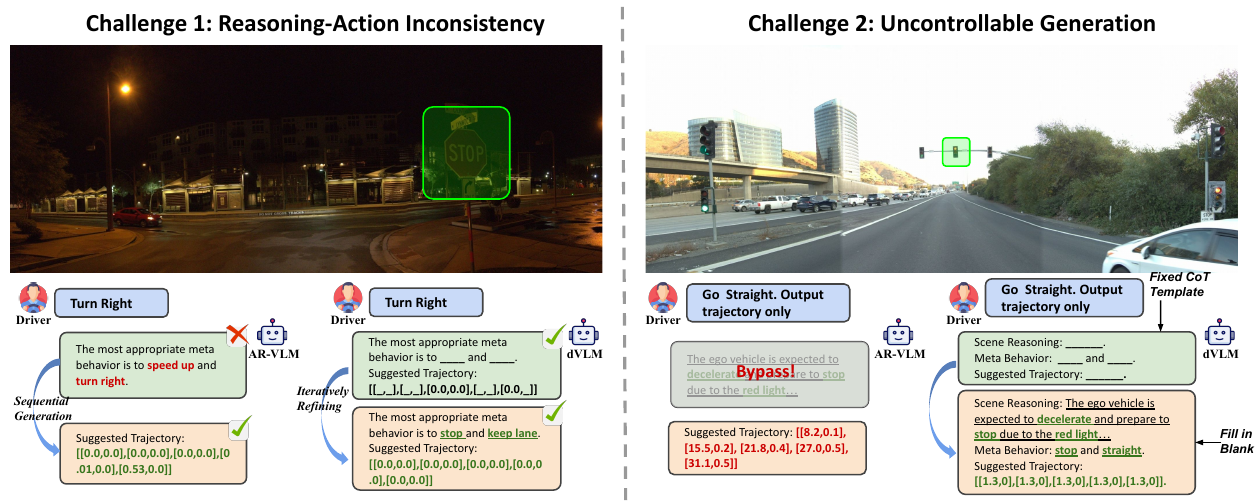}
    \vspace{-3mm}
    \caption{Challenges in existing driving VLMs/VLAs. (1) Reasoning–action inconsistency: ARM-based VLMs’ sequential decoding can produce plans where the inferred behavior conflicts with the predicted trajectory (e.g., behavior mismatches the trajectory). Our dVLM performs template-anchored iterative refinement with bidirectional attention, enforcing cross-field consistency between meta-behavior and trajectory. (2) Uncontrollable generation: The structured reasoning of AR-VLMs is easily corrupted by prompt-level perturbations (e.g., bypassing reasoning steps), leading to broken formats and unstable outputs. In contrast, dVLM’s template-anchored fill-in-the-blank decoding with schema checks preserves order and semantics, yielding safe, consistent actions.}
    \label{fig:challenge}
    \vspace{-5mm}
\end{figure*}

End-to-end (E2E) autonomous driving systems have achieved remarkable progress by unifying perception, prediction, and planning within a single framework~\cite{jiang2023vad,hu2023planning}. Building on this trend, increasing attention has been devoted to enhancing the robustness of E2E systems in long-tail and out-of-distribution scenarios~\cite{tian2024tokenize,tian2024drivevlm,hwang2024emma,xu2024vlm}. Motivated by this goal, recent efforts explore the integration of vision-language-action (VLA) models~\cite{black2024pi0visionlanguageactionflowmodel, intelligence2025pi05visionlanguageactionmodelopenworld, brohan2023rt2visionlanguageactionmodelstransfer,kim2024openvlaopensourcevisionlanguageactionmodel}, which leverage rich world knowledge and strong reasoning capabilities to advance driving intelligence~\cite{zhou2025opendrivevla,zhou2025autovla,chi2025impromptu,jiang2025irl,arai2025covla}. This integration not only enables driving systems to reason over open-vocabulary visual cues and follow natural-language navigation commands, but also enhances interpretability of driving systems - an essential property for ensuring safe and reliable decision-making in complex, real-world driving environments. 
However, existing mainstream VLMs/VLAs are predominantly built upon autoregressive models (ARMs)~\cite{alayrac2022flamingo,
liu2024improved,
liu2023visual,
li2024llava}. These models typically take camera frames and textual instructions as inputs and then sequentially generate the reasoning process followed by action tokens, owing to their inherently unidirectional (causal) attention mechanism. Despite remarkable advances, a fundamental question remains unresolved: \emph{Is the autoregressive paradigm truly the optimal paradigm for VLM-based E2E autonomous driving system?}

In this paper, we observe two critical limitations of ARM-based VLMs/VLAs for autonomous driving:
(1) \textbf{Reasoning–action inconsistency~\cite{nvidia2025alpamayor1bridgingreasoningaction}:} As shown in Figure~\ref{fig:comparison}, during generation, ARM-based VLMs/VLAs can only attend to previously generated tokens but not subsequent action tokens. As a result, the model may produce actions that are semantically misaligned with its reasoning, leading to inconsistent or unsafe decisions~(see Figure~\ref{fig:challenge} for an example). 
(2) \textbf{Uncontrollable generation:}
ARMs rely solely on likelihood optimization and format-based rewards~\cite{guo2025deepseek}, without ensuring adherence to a standardized reasoning process, which embeds critical safety constraints. As a result, their reasoning process is highly sensitive to input instructions~\cite{kuo2025h,huang2025pathdriftlargereasoning,robey2025jailbreaking}. 
For instance, as illustrated in Figure~\ref{fig:challenge}, the driver can easily gain unintended control over the model by issuing a simple instruction such as ``predict the trajectory without the reasoning process''. This instruction causes the ARM to bypass its internal structured reasoning pipeline (e.g., recognizing a red light and decelerating) and directly generate an unsafe trajectory.


To mitigate these limitations, we go beyond auto-regressive decoding to diffusion-based VLMs~(dVLMs)~\cite{you2025llada,yu2025dimple,yang2025mmada}. The diffusion-based paradigms~\cite{ye2025dream,nie2025large,deepmind2025geminidiffusion} iteratively denoise a heavily corrupted sequence, allowing for bidirectional context integration in each refinement step instead of left-to-right causal decoding. This architectural shift directly addresses ARM-based exposure-bias hallucination~\cite{huang2025survey} by letting every position condition on global context throughout generation, which empirically yields more coherent, globally consistent outputs~\cite{li2022diffusion}. Just as importantly, the multi-step denoising process naturally supports controllable generation: safety constraints, structured templates, or partial tokens can steer the denoising trajectory without retraining, enabling stronger adherence to prescribed reasoning protocols and reducing susceptibility to prompt manipulation~\cite{gulrajani2023likelihood,li2022diffusion,xiong2025unveiling}. These properties - bidirectional reasoning, global consistency, and controllability - make dVLMs a principled alternative for safety-critical driving systems. \looseness=-1

Inspired by these insights, to address the limitation of ARMs for autonomous driving, we propose \model{}, a diffusion-based vision–language model for driving. 
Our key idea is to replace left-to-right decoding with bidirectional, iterative generation, allowing visual understanding, textual reasoning, and low-level trajectory tokens to be refined jointly, thereby achieving stronger global consistency and controllability. In Figure~\ref{fig:challenge}, \model{} employs a structured CoT template with fill-in-the-blank so that the reasoning is anchored. This design preserves consistency between reasoning and action while preventing users from easily exerting unintended control over the model, thereby ensuring that it predicts safe behaviors (e.g., ``decelerate and stop'') through the structured reasoning process.

To implement it, the most straightforward way is to train the diffusion model on the structured reasoning-action data for the driving domain. Despite its simplicity, such naive training is not effective. 
We find that naive training will introduce a slot-length bias, where reasoning steps, such as scene understanding, behavior prediction, and trajectory generation, naturally vary in length, yet controllable decoding requires predefined slot sizes at inference. Once these slots are fixed, the model tends to ``fill to capacity'', generating outputs constrained by length rather than semantics. 
For example, when a window size of three tokens is reserved for meta-behavior prediction, the model may incorrectly produce \texttt{right-lane-change} instead of the correct \texttt{keep-lane}, simply because the longer phrase fits the allocated token window. 
To address this challenge, we introduce a dynamic denoising strategy that allows variable-length infilling within a fixed mask window. 
Built upon this decoding strategy, \model{} is adapted to the autonomous driving domain based on the LLaDA-V~\cite{you2025llada} through a two-stage training pipeline: (i) large-scale alignment using 145k driving-related QA pairs curated from open-source datasets~\cite{chi2025impromptu, hao2025driveaction,sima2024drivelm,wang2024drivecot,arai2025covla} to ground the model’s multimodal understanding in the driving domain; and (ii) supervised fine-tuning with 53k \emph{structured reasoning-action} annotations for driving. 
After training, \model{} preserves the controllability of diffusion-based decoding while mitigating slot-length bias, leading to reasoning and planning outputs that are semantically grounded and behaviorally consistent for autonomous driving domain. 

We evaluate the effectiveness of \model{} on the Waymo Open Dataset (WOD-E2E) and nuScenes~\cite{waymo_e2e_2025,caesar2020nuscenes}. To assess reasoning–action consistency, we compare \model{} against an autoregressive (ARM)-based vision–language agent (VLA) under identical training settings. Our method demonstrates a significant improvement over ARM-based counterparts. 
In terms of driving performance,  when trained on the same data, our diffusion-based approach surpasses ARM-based baselines with a 6\% improvement in long-tail scenarios on WOD-E2E.
Despite being built upon LLaDA-V, a relatively modest backbone compared to Qwen2.5-VL and Gemini, \model{}  can also achieve comparable driving performance to these VLM/VLA systems that are built on stronger base models such as the Qwen and Gemini family~\cite{qwen3technicalreport}.
Furthermore, we analyze the effectiveness of the proposed dynamic denoising strategy and perform instruction perturbation experiments to validate its critical role in maintaining controllable reasoning and ensuring safe, reliable planning behaviors.

\section{Related Work}

\begin{figure*}[!t]
    \centering
    \includegraphics[width=1.0\linewidth,
    trim=0 0pt 0 0pt, 
    clip]{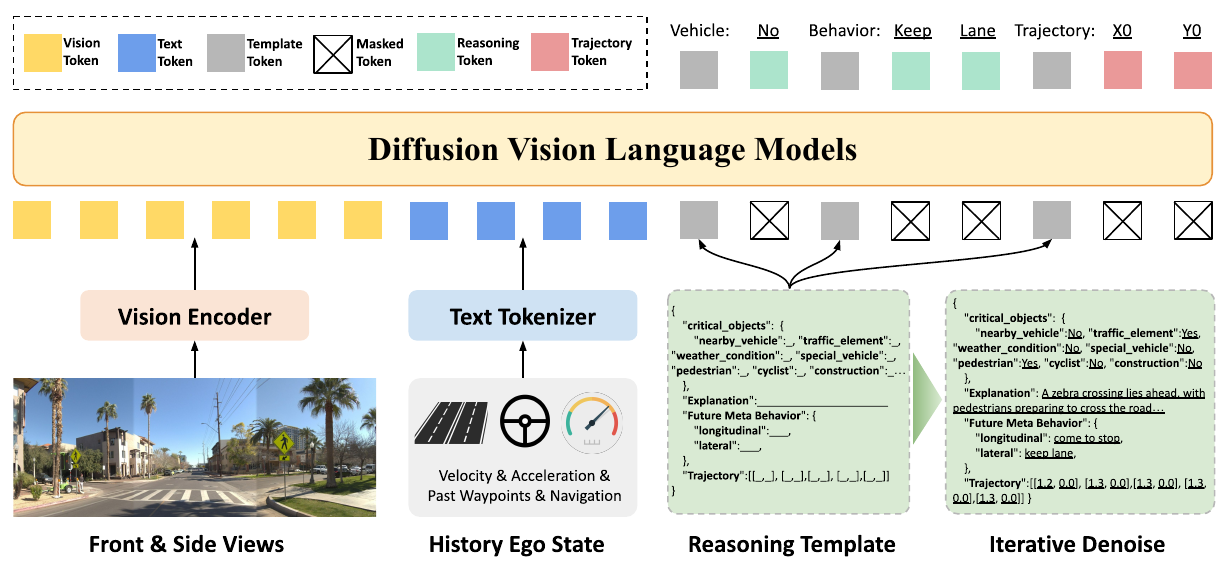}
    \vspace{-4mm}
    \caption{The overview of \model{} framework.}
    \label{fig:framework}
    \vspace{-4mm}
\end{figure*}

\subsection{Vision-Language-Action~(VLA) Models.}
Recent progress in Vision–Language–Action (VLA) models has unified perception, reasoning, and control within a single multimodal framework. Most existing VLAs are autoregressive, representing actions as discrete tokens decoded sequentially by large language models. Systems such as RT-2~\cite{brohan2023rt2visionlanguageactionmodelstransfer}, OpenVLA~\cite{kim2024openvlaopensourcevisionlanguageactionmodel}, and their driving counterparts—OpenDriveVLA~\cite{zhou2025opendrivevlaendtoendautonomousdriving}, Impromptu-VLA~\cite{chi2025impromptuvlaopenweights}, and AutoVLA~\cite{zhou2025autovlavisionlanguageactionmodelendtoend}—show that leveraging language-driven reasoning improves trajectory prediction, particularly in long-tail scenarios. More recently, diffusion- or flow-based VLAs (e.g., Octo~\cite{octomodelteam2024octoopensourcegeneralistrobot}, $\pi_0$~\cite{black2024pi0visionlanguageactionflowmodel}, $\pi_{0.5}$~\cite{intelligence2025pi05visionlanguageactionmodelopenworld}, DiffVLA~\cite{jiang2025diffvla}) generate continuous trajectories through iterative denoising or flow matching, enhancing open-world robustness. Other efforts such as LoHoVLA~\cite{yang2025lohovlaunifiedvisionlanguageactionmodel}, ChatVLA-2~\cite{zhou2025chatvla2visionlanguageactionmodelopenworld}, CogVLA~\cite{li2025cogvlacognitionalignedvisionlanguageactionmodel}, and NORA~\cite{hung2025norasmallopensourcedgeneralist} focus on long-horizon reasoning and efficiency. While these approaches advance generalization and capability, challenges remain in ensuring reasoning–action consistency and safe, controllable decision-making—gaps that motivate our diffusion-based formulation in autonomous driving. \looseness=-1

\vspace{-2mm}
\subsection{Diffusion Large Language Models (dLLMs).}

Discrete diffusion modeling has advanced rapidly in both theory and practice. Early works such as D3PMs formalized noise and transition designs~\citep{austin2021structured}, while Diffusion-LM showed controllable text generation beyond the autoregressive paradigm~\citep{li2022diffusion}. Subsequent studies refined discrete objectives and simplified masked diffusion for better stability and efficiency~\citep{lou2024discrete, sahoo2024simple, shi2024simplified}.
Building on these foundations, dLLMs have scaled to large language model regimes: the LLaDA series~\cite{nie2025large} and Dream 7B~\cite{ye2025dream} match autoregressive performance via large-scale masked diffusion training and careful scheduling~\citep{nie2025large}, while post-training methods such as LLaDA 1.5 and TraceRL use preference optimization and trajectory-aware RL to better align with human feedback and reasoning goals~\citep{zhu2025llada15variancereducedpreference, wang2025revolutionizingreinforcementlearningframework}.
Closer to our multimodal setting, MMaDA~\cite{yang2025mmada} proposes a unified, modality-agnostic diffusion architecture with chain-of-thought alignment and unified RL post-training; LLaDA-V~\cite{you2025llada} and Dimple~\cite{yu2025dimple} integrate visual instruction tuning into masked diffusion LMs and report competitive multimodal performance; and some recent works~\cite{wen2025llada,liang2025discrete,li2025discrete,wen2025dvla} extend diffusion VLMs to vision–language–action for embodied AI systems. Along this path, to the best of our knowledge, our work is the first to introduce a reasoning-based dVLM for autonomous driving. \looseness=-1

\section{Preliminaries}

\subsection{CoT-based Planning in Autonomous Driving}

In autonomous driving, reasoning-based planning aims to generate interpretable motion conditioned on perception and intention~\cite{luo2025adathinkdrive,
yuan2025autodrive,
zeng2025futuresightdrive,
li2025recogdrive,
rowe2025poutine}. 
Given past multi-camera frames \(v_{T-K+1:T}\), historical ego states \(e_{T-K+1:T}\), and a navigation command \(g\), the planner outputs a Chain-of-Thought~(CoT) reasoning process \(R\) and a sequence of future waypoints \(\boldsymbol{\tau}=\{(x^i,y^i)\}_{i=1}^{K}\). 
Formally, the reasoning planner can be represented as
\vspace{-3mm}
\begin{equation}
(R, \boldsymbol{\tau}) = \mathrm{Planner}(v_{T-K+1:T}, e_{T-K+1:T}, g),
\end{equation}
where $R$ captures the intermediate reasoning steps that bridge perception and decision, and $\boldsymbol{\tau}$ defines the ego-centric trajectory to be executed. 
This formulation unifies perception, reasoning, and motion generation within an interpretable framework, enabling safer and more transparent planning in complex driving environments.

\subsection{Discrete Diffusion Vision Language Models} \label{sec:dllm}

\paragraph{Forward and Reverse Process.}
Let $\mathbf{x}_0=(x_1,\ldots,x_L)$ be the target token sequence and $\mathbf{c}=(\mathbf{v},\mathbf{p})$ the conditioning context (visual features and text prompt).
A discrete diffusion VLM follows a \emph{mask--denoise} cycle governed by a noise schedule $\{\lambda_t\}_{t=1}^{T}$.
At step $t$, a subset $\mathcal{M}_t\!\subseteq\!\{1{:}L\}$ is masked according to $\lambda_t$, yielding a corrupted sequence $\mathbf{x}_t$.
The reverse process applies a denoising policy $\pi_{\theta}$ that, given $(\mathbf{x}_t,\mathbf{c},t)$, predicts replacements only for indices in $\mathcal{M}_t$ while keeping visible tokens fixed; iterating $t{=}T\!\to\!0$ recovers a clean sequence.
This view matches absorbing-state discrete diffusion with categorical transitions and naturally supports bidirectional conditioning on $\mathbf{c}$ and $\mathbf{x}_t$.

\vspace{-4mm}
\paragraph{Discrete Diffusion Modeling.}
Modeling operates directly in token space with a special $[\mathrm{MASK}]$ state.
Training samples $t$ and masked sets $\mathcal{M}_t$, then optimizes $\theta$ so that $\pi_{\theta}$ (equivalently the denoiser $p_{\theta}$) predicts ground-truth tokens $\{x_i\}_{i\in\mathcal{M}_t}$ conditioned on $(\mathbf{x}_t,\mathbf{c})$; the loss is computed only on masked positions:
\begin{equation}
\label{eq:loss1}
\mathcal{L}(\theta)=
\mathbb{E}_{t,\mathbf{x}_0,\mathbf{x}_t}\!\left[
-\frac{1}{|\mathcal{M}_t|}\sum_{i\in\mathcal{M}_t}
\log p_{\theta}\!\big(x_0^{i}\mid \mathbf{x}_t,\mathbf{c}\big)
\right],
\end{equation}
where $\mathbf{x}_t$ is obtained by masking $\mathbf{x}_0$ at step $t$ and $\mathcal{M}_t=\{i:\,x_t^{i}=[\mathrm{MASK}]\}$.

\section{Method}

In this section, we introduce \model{}, a diffusion-based vision–language model for end-to-end autonomous driving that unifies visual perception, object detection, high-level prediction, and low-level planning within a single reasoning pipeline. We first present our framework, then formalize controllable reasoning for driving and instantiate it with a dynamic denoise strategy that progressively reveals masked targets during denoising ($\S$~\ref{sec:framework}). Finally, we describe dataset construction, detailing data sources and the annotation protocol that supervises each reasoning stage~($\S$~\ref{sec:data}). \looseness=-1

\subsection{The \model{} Framework} \label{sec:framework}

\paragraph{Model Architecture.}
The overall architecture of \model{} is illustrated in Figure~\ref{fig:framework}. \model{} is initialized from LLaDA-V~\cite{you2025llada}, a diffusion vision language model whose multimodal understanding is comparable to open-source autoregressive VLMs~\cite{li2024llava,wang2024qwen2,bai2025qwen2}. The \model{} comprises three components: (i) a LLM backbone, (ii) a vision encoder, and (iii) a multimodal projector. We adopt LLaDA-8B-Instruct~\cite{nie2025large} as the language backbone, SigLIP2-so400m-patch14-384~\cite{tschannen2025siglip} as the vision encoder, and a lightweight MLP as the projector. \looseness=-1

\model{} consumes camera inputs $\mathcal{V}$, high\mbox{-}level navigation instructions \(g\), and ego\mbox{-}state signals \(e\), and produces scene reasoning and a planned trajectory. We tailor  $\mathcal{V}$ to each dataset: for \textbf{nuScenes}~\cite{caesar2020nuscenes}, we use the \emph{front} view from the past one second,
\[
\mathcal{V}_{\text{nuScenes}}=\big[c_{\text{front}}^{\,t-1\text{s}},\;c_{\text{front}}^{\,t-0.5\text{s}},\;c_{\text{front}}^{\,t}\big],
\]
providing short\mbox{-}term temporal context; for \textbf{Waymo Open Dataset End-to-End}~\cite{xu2025wod}, we use three synchronized views at the current time step,
\[
\mathcal{V}_{\text{Waymo}}=\big\{c_{\text{front-left}}^{\,t},\;c_{\text{front}}^{\,t},\;c_{\text{front-right}}^{\,t}\big\}.
\]
The navigation \(g\) (e.g., \emph{Turn Left}, \emph{Go Straight}) specifies the intended maneuver. The ego state \(e\) aggregates signals over the past 3\,s at 0.5\,s intervals (2\,Hz), including current velocity, acceleration, and a short history of recent waypoints. Given these inputs, \model{} first generates a structured, controllable reasoning trace and then predicts the vehicle’s future behaviors at 2\,Hz.

\vspace{-5mm}
\paragraph{Action Representation.} There are two prevailing action representations for VLM-based planning: textual waypoints and discrete action tokens. Textual waypoints serialize the future path as a schema-constrained string (e.g., JSON coordinates)~\cite{tian2024drivevlm, chi2025impromptu, rowe2025poutine}, while action-token methods emit discrete action/trajectory tokens that an expert head dequantizes into waypoints~\cite{zhou2025autovla, zhou2025opendrivevla}. Across standard benchmarks, both families deliver comparable planning quality. Action tokens are appealing for their lower decoding latency and for easing the enforcement of physical/dynamics priors via dequantization and expert decoders. Nevertheless, we still adopt textual waypoints for two reasons: (1) a unified language interface--perception, prediction, and coordinates in one channel; and (2) diffusion language-model fine-tuning is notably unstable in practice~\cite{you2025llada}, and introducing extra action vocabularies/decoder heads tends to amplify this instability and data demand in our setting.
Specifically, \model{} predicts $\boldsymbol{\tau}=\{(x^i,y^i)\}_{i=1}^{K}$ at $2\,\mathrm{Hz}$. Notably, unlike autoregressive models (ARMs), \model{} can dynamically adjust the number $K$ of generated waypoints at inference by setting the number of mask tokens, and is therefore not constrained by the fixed trajectory length used during training.

\vspace{-5mm}
\paragraph{Controllable Reasoning for Driving.} Recent studies emphasize that the reasoning process in autonomous driving should be \emph{structured}, reflecting the inherently sequential pipeline of driving: perception, prediction, and planning~\cite{sima2024drivelm, tian2024drivevlm}. However, existing ARM-based LLMs exhibit unreliability when required to generate structured output. Inspired by prior work~\cite{xiong2025unveiling}, we pioneer the use of diffusion vision–language models to produce \emph{structured, controllable} reasoning and actions in autonomous driving. Our key idea is to recast free-form generation as \emph{constrained infilling}: we first construct a structured reasoning template $\hat{\mathbf{x}}_{T}$, as shown in Figure~\ref{fig:framework}. Unlike standard dVLM decoding that begins from a fully masked sequence $\mathbf{x}_{T}$ (i.e., $\mathcal{M}_{T}=\{1{:}L\}$), we initialize the reverse chain with $\hat{\mathbf{x}}_{T}$ so that only a subset of slots is masked, $\mathcal{M}_{T}\subsetneq\{1{:}L\}$. The visible anchors in $\hat{\mathbf{x}}_{T}$ act as hard constraints and guidance, and the denoising policy $\pi_{\theta}$ fills the remaining slots in a controlled, template-consistent manner. Formally, our \emph{structured} objective conditions on a template with visible anchors.
Let $\hat{\mathbf{x}}_{T}$ be the template, $\mathcal{A}=\{i:\hat{x}_{T}^{i}\neq[\mathrm{MASK}]\}$ the anchor set,
and $\mathcal{E}=\{1{:}L\}\setminus\mathcal{A}$ the editable indices.
At step $t$, we mask only editable slots $\mathcal{M}_t\subseteq\mathcal{E}$ and keep anchors fixed in $\mathbf{x}_t$.
The training loss is:
\begin{equation}
\label{eq:loss2}
\mathcal{L}(\theta)=
\mathbb{E}_{t,\mathbf{x}_0,\mathbf{x}_t,\;\mathcal{M}_t\subseteq\mathcal{E}}
\!\left[
-\frac{1}{|\mathcal{M}_t|}\sum_{i\in\mathcal{M}_t}
\log p_{\theta}\!\big(x_0^{\,i}\mid \mathbf{x}_t,\mathbf{c}\big)
\right],
\end{equation}
where $\mathbf{x}_t$ is obtained by masking $\mathbf{x}_0$ only on $\mathcal{M}_t$ and clamping $x_t^{i}=\hat{x}_{T}^{i}$ for all $i\in\mathcal{A}$.

\begin{figure}[!t]
    \centering
    \includegraphics[width=1.0\linewidth]{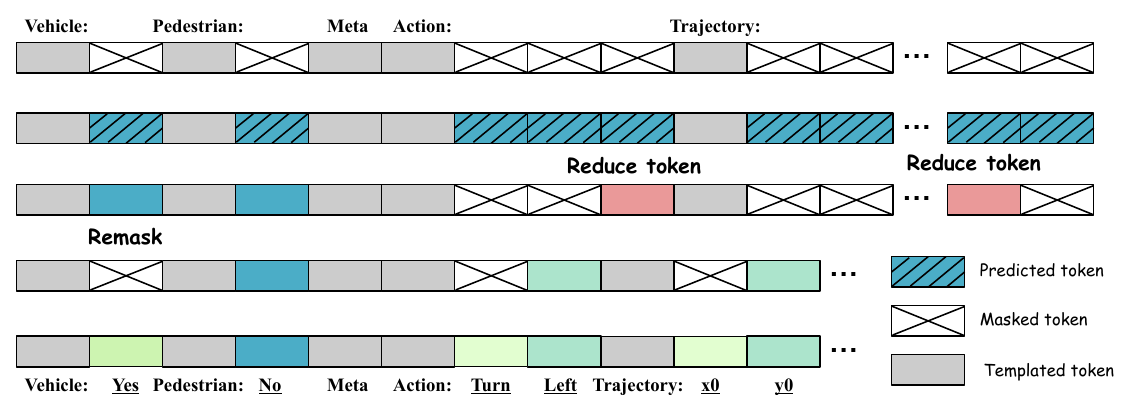}
    \vspace{-6mm}
    \caption{Dynamic denoise strategy for controllable reasoning.}
    \label{fig:denoise}
    \vspace{-5mm}
\end{figure}

\begin{table*}[!t]
\centering
\caption{Evaluation of Object$\leftrightarrow$Explanation consistency and Behavior$\leftrightarrow$Trajectory alignment on nuScenes and WOD-E2E validation sets. \textbf{Col./RFS} denotes the average \textbf{Collision Rate (Col.)}—lower is better—and the \textbf{Rater Feedback Score (RFS)}—higher is better.}
\vspace{-3mm}
\small
\resizebox{0.95\textwidth}{!}{
\begin{tabular}{
    >{\centering\arraybackslash}p{2.4cm}  
    >{\centering\arraybackslash}p{1.2cm}  
    >{\centering\arraybackslash}p{1.7cm}  
    >{\centering\arraybackslash}p{1.2cm}  
    >{\centering\arraybackslash}p{1.2cm}  
    >{\centering\arraybackslash}p{1.2cm}  
    >{\centering\arraybackslash}p{1.2cm}  
    >{\centering\arraybackslash}p{1.2cm}  
    >{\centering\arraybackslash}p{1.2cm}  
    >{\centering\arraybackslash}p{1.2cm}  
}
\toprule
\multirow{2}{*}{\textbf{Method}} &
\multirow{2}{*}{\makecell[c]{\textbf{Decoding} \\ \textbf{Strategy}}} &
\multirow{2}{*}{\textbf{Backbone}} &
\multicolumn{3}{c}{\textbf{Object$\leftrightarrow$Explanation}} &
\multicolumn{3}{c}{\textbf{Behavior$\leftrightarrow$Trajectory}} &
\multirow{2}{*}{\textbf{Col./RFS}} \\
\cmidrule(lr){4-6} \cmidrule(lr){7-9}
& & & \textbf{O} $\rightarrow$ \textbf{E} & \textbf{E} $\rightarrow$ \textbf{O} & \textbf{Avg.} & \textbf{longitudinal} & \textbf{lateral} & \textbf{Avg.} & \\
\midrule
\multicolumn{10}{c}{\textit{nuScenes validation set}} \\
\midrule
OpenEMMA$^{\ast}$~\cite{xing2025openemmaopensourcemultimodalmodel} & AR & InternVL3.5 & 89.5 & 84.8 & 87.2 & 67.6 & 79.6 & 73.6 & -- \\
LightEMMA$^{\ast}$~\cite{qiao2025lightemma} & AR & Qwen3-VL & 87.7 & 85.8 & 86.7 & 76.1 & \underline{87.2} & 81.7 & -- \\
VLM-AD~(ours) & AR & LLaVA-OV & \underline{93.2} & \underline{96.3} & \underline{94.7} & \underline{85.3} & 85.7 & \underline{85.5} & \underline{0.35} \\
\model{}~(ours) & Diff. & LLaDA-V & \textbf{98.2} & \textbf{98.9} & \textbf{98.6} & \textbf{87.1} & \textbf{88.5} & \textbf{87.8} & \textbf{0.32} \\
\midrule
\multicolumn{10}{c}{\textit{WOD-E2E validation set}} \\
\midrule
OpenEMMA$^{\ast}$~\cite{xing2025openemmaopensourcemultimodalmodel}    & AR        & InternVL3.5 & 89.4 & 85.1  & 87.3 & 44.2 & 39.6 & 41.9 & 5.158 \\
LightEMMA$^{\ast}$~\cite{qiao2025lightemma}      & AR        & Qwen3-VL & 87.8  & 86.0 & 86.9 & \underline{71.3} & \underline{71.6} & \underline{71.5} & 6.517 \\
VLM-AD~(ours)      & AR        & LLaVA-OV & \underline{91.8} & \underline{93.7} & \underline{92.8} & 59.7 & 85.9 & 72.8 & \underline{7.215} \\
\model{}~(ours)       & Diff. & LLaDA-V & \textbf{98.1} & \textbf{98.3} & \textbf{98.2} & \textbf{74.4} & \textbf{85.7} & \textbf{80.1} & \textbf{7.633} \\
\bottomrule
\end{tabular}}
\label{tab:eval_consistency}
\vspace{-5mm}
\end{table*}

\vspace{-5mm}
\paragraph{Dynamic Denoise Strategy.} \label{sec:denoise}

As discussed in Section~\ref{sec:dllm}, diffusion-based LLMs require specifying a fixed-length masked token sequence before the reverse denoising process. Although several recent studies have attempted to relax this constraint by enabling variable-length masking~\cite{li2025beyond, li2025diffusion}, the problem remains unresolved. Specifically, while structured templates offer a means of enforcing controllable reasoning, their dependence on fixed-length slots remains inherently restrictive. Owing to lexical variability, tokenized phrases exhibit inconsistent lengths---for example, ``left turn'' may consist of only two subword tokens, whereas ``right lane change'' typically requires three or more. A naive strategy is to allocate slots according to the maximum span length, yet this introduces a \emph{length-matching bias}: the model is implicitly incentivized to generate content whose length merely fits the available blanks rather than content that is semantically appropriate (e.g., ``three blanks $\Rightarrow$ three-token phrase''). Therefore, we propose a \textbf{dynamic denoise strategy} for controllable reasoning, as illustrated in Figure~\ref{fig:denoise}. The key idea is to introduce a special \textit{reduce token} that allows the masked span to be adaptively shortened during the reverse denoising process. At initialization, all non-templated positions between two structural anchors are fully masked. During each denoising step, the model predicts token distributions over these masked slots. Once the confidence of placing a reduce token at a specific position exceeds a predefined threshold, that position is \emph{fixed} as a reduce token, and all masked tokens following it—up to the next templated token—are immediately pruned from the sequence.
This reduction is only valid under a structural constraint: every token between the reduced token and the nearest subsequent templated token must still be masked at the moment of reduction. If any of these positions have already been filled with non-masked content, the reduction is rejected to preserve alignment with the template. In this way, the model is permitted to terminate generation early only when the remaining span is semantically redundant, rather than being forced to fill a pre-allocated slot length. Notably, we observe in practice that the model tends to assign the reduce token prematurely. To avoid this, we employ a brief \textit{warm-up} phase during which the \textit{reduce token} is not fixed, allowing the model to first decode meaningful content before reduction is activated.

\subsection{Dataset Construction}\label{sec:data}
Unlike contemporary state-of-the-art VLMs (e.g., Qwen3-VL~\cite{qwen3technicalreport}, InternVL~3.5~\cite{wang2025internvl3}), our foundation dVLM, LLaDA-V, has not been extensively pre-trained on massive multimodal corpora. As a result, its world knowledge—particularly for driving—lags behind mainstream ARM-based VLMs. To align LLaDA-V with driving-specific knowledge, we curate and process data from existing driving datasets, focusing on supervision signals that emphasize scene understanding, high-level prediction, and planning.
Concretely, we collect and filter examples from \emph{ImpromptuVLA}~\cite{chi2025impromptu}, \emph{DriveAction}~\cite{hao2025driveaction}, \emph{DriveLM}~\cite{sima2024drivelm}, \emph{DriveLMMo1}~\cite{ishaq2025drivelmmo1stepbystepreasoningdataset}, and \emph{CoVLA}~\cite{arai2025covla} and obtain approximately \textbf{145k} driving-related QA pairs for alignment, providing targeted supervision that strengthens LLaDA-V's perception and prediction capabilities. Details can be found in Appendix~A.
Moreover, high–quality reasoning signals are essential for linking scene semantics to the ego vehicle’s future motion~\cite{zhou2025autovla,nvidia2025alpamayor1bridgingreasoningaction}. We therefore build a structured annotation pipeline tailored to trajectory prediction. Specifically, we construct reasoning annotations tailored to trajectory prediction by employing GPT-4.1~\cite{openai_gpt41_2025} as an automatic annotator conditioned on the same signals available to the predictor—multi-frame front-view images, past ego state, a navigation command, and the ego’s future waypoints—and by overlaying 2D bounding boxes on the current frame to strengthen object grounding. Inspired by Poutine~\cite{rowe2025poutine}, we structure the output into four fields where \texttt{object detection} lists salient agents and traffic elements, \texttt{explanation} provides a concise causal rationale linking those cues to the upcoming motion, \texttt{future behavior} specifies high-level longitudinal and lateral intentions consistent with the scene, and \texttt{trajectory} encodes the target future path as waypoints. At inference time, the dVLM treats these four fields as templated slots and fills the corresponding values. Finally, we annotate \textbf{~23k} and \textbf{~30k} reasoning data for nuScenes and WOD-E2E datasets, respectively. \looseness=-1

\section{Experiment}

\begin{table*}[!t]
\caption{Comparison of different methods on WOD-E2E test set. $^{\ast}$ indicates models evaluated in a zero-shot setting (without fine-tuning).}
\vspace{-3mm}
\label{tab:waymo}
\centering
\resizebox{0.84\textwidth}{!}{
\begin{tabular}{p{2.7cm}<{\centering} p{2.2cm}<{\centering} p{2.2cm}<{\centering} p{1.8cm}<{\centering} p{1.8cm}<{\centering} p{1.8cm}<{\centering}}
\toprule
\textbf{Method} & \textbf{Backbone} & \textbf{\# Trajectories} & \textbf{RFS}$\uparrow$ & \textbf{ADE (5s)}$\downarrow$ & \textbf{ADE (3s)}$\downarrow$ \\
\midrule
OpenEMMA$^{\ast}$~\cite{xing2025openemmaopensourcemultimodalmodel}  & -        & zero-shot    & 5.158 & 12.476 & 6.684 \\
LightEMMA$^{\ast}$~\cite{qiao2025lightemma}  & -        & zero-shot     & 6.517 & 3.740 & 1.705 \\
Open-LLaMA~\cite{waymo_e2e_2025} & LLaMA-Vision        & --     & 7.429 & 3.217 & \underline{1.314} \\
NaiveEMMA~\cite{waymo_e2e_2025} & Gemini        & --     & 7.528 & 3.018 & 1.320 \\
AutoVLA~\cite{zhou2025autovla}   & Qwen2.5-VL    & 52.8k  & \underline{7.557} & \textbf{2.558} & 1.351 \\
\midrule
dVLM-AD   & LLaDA-V       & 29.3k  & \textbf{7.633} & 3.022 & \textbf{1.285} \\
\bottomrule
\end{tabular}}
\vspace{-5mm}
\end{table*}

\begin{table}[t]
\centering
\caption{Comparison of driving-related models on L2 Error (m) across different horizons on nuScenes validation dataset. $^{\ast}$ indicates models evaluated in a zero-shot setting (without fine-tuning).}
\vspace{-3mm}
\resizebox{0.41\textwidth}{!}{
\begin{tabular}{lcccc}
\toprule
\multirow{2}{*}{\textbf{Method}}  & \multicolumn{4}{c}{\textbf{L2 Error (m) $\downarrow$}} \\
\cmidrule(lr){2-5}
 & 1s & 2s & 3s & \cellcolor[HTML]{E8E8FF}\textbf{Avg.} \\
\midrule
\multicolumn{5}{l}{\textit{Training-based Driving Policy}} \\
UniAD~\cite{hu2023planning}         & 0.20 & 0.42 & 0.75 & \cellcolor[HTML]{E8E8FF}0.46 \\
VAD-Base~\cite{jiang2023vad}      & 0.17 & 0.34 & 0.60 & \cellcolor[HTML]{E8E8FF}0.37 \\
BEV-Planner~\cite{li2024ego}   & 0.16 & 0.32 & 0.57 & \cellcolor[HTML]{E8E8FF}0.35 \\
Ego-MLP~\cite{li2024ego}      & 0.15 & 0.32 & 0.59 & \cellcolor[HTML]{E8E8FF}0.35 \\
\midrule
\multicolumn{5}{l}{\textit{VLMs or VLAs with Reasoning}} \\
OpenEMMA$^{\ast}$~\cite{chi2025impromptu}       & 1.45 & 3.21 & 3.76 & \cellcolor[HTML]{E8E8FF}2.81 \\
Gemini-2.5-Pro$^{\ast}$~\cite{chi2025impromptu}  & 0.37 & 1.35 & 2.96 & \cellcolor[HTML]{E8E8FF}1.56 \\
Claude-3.7-Sonnet$^{\ast}$~\cite{chi2025impromptu}  & 0.28 & 0.94 & 2.04 & \cellcolor[HTML]{E8E8FF}1.09 \\
GPT-4o$^{\ast}$~\cite{chi2025impromptu}        & 0.28 & 0.93 & 2.02 & \cellcolor[HTML]{E8E8FF}1.07 \\
DriveVLM~\cite{tian2024drivevlm}  & 0.18 & \textbf{0.34} & \underline{0.68} &   \cellcolor[HTML]{E8E8FF}\textbf{0.40} \\
AutoVLA (w/ CoT)~\cite{zhou2025autovla} & 0.25 & 0.46 & 0.73 & \cellcolor[HTML]{E8E8FF}0.48 \\
\midrule
\textbf{\model{}} (ours) & \textbf{0.15} & \underline{0.40} & \textbf{0.68} & \cellcolor[HTML]{E8E8FF}\underline{0.41} \\
\bottomrule
\end{tabular}}
\label{tab:nuscenes}
\vspace{-5mm}
\end{table}

\subsection{Experimental Setup}

\paragraph{Datasets.}
We evaluate VLM-AD and \model{} in nuScenes~\cite{caesar2020nuscenes} and WOD-E2E~\cite{xu2025wod} benchmarks in open-loop settings. \textbf{nuScenes} contains 1{,}000 urban driving scenes (20\,s each) captured in Boston and Singapore with a full AV sensor suite (6 cameras, 5 radars, 1 lidar). The official split is 700/150/150 scenes for train/val/test. Camera frames are high-resolution \(1600\times900\) JPEGs recorded at \(\sim12\) Hz, while annotated keyframes are sampled at 2 Hz. 
\textbf{Waymo Open Dataset End-to-End (WOD-E2E)} comprises 4{,}021 long-tail driving segments of 20\,s each, split into 2{,}037/479/1{,}505 for train/val/test. Each segment includes eight-camera surround-views (\(1920\times1280\)) video at 10 Hz along with routing inputs and ego state. For evaluation, only the first 12\,s of each test segment are provided; the subsequent 8\,s future is hidden and predictions must be generated using information available up to the last frame of the 12\,s context. \looseness=-1

\vspace{-4mm}
\paragraph{Implementation Details.}
For a fair comparison, we train both VLM-AD and \model{} with the same two-stage recipe: in Stage I, we optimize 145k alignment samples with Eq.~\ref{eq:loss1} for 1 epoch; in Stage II, we finetune the models with Eq.~\ref{eq:loss2} for 3 epochs on nuScenes (23k) and WOD-E2E (30k). Details are provided in the appendix.

\vspace{-4mm}
\paragraph{Evaluation Metrics.}
For nuScenes, we report the L2 distance error computed on the official 2\,Hz keyframes.
For WOD-E2E, we report ADE and RFS. ADE is the mean Euclidean distance over the prediction horizon, $\frac{1}{H}\sum_{t=1}^{H}\lVert \hat{y}_t - y_t\rVert_2$, averaged over samples. 
RFS (Rater Feedback Score)~\cite{xu2025wod} evaluates a single predicted ego trajectory against multiple human-rated reference trajectories using trust regions. For each rater trajectory, rectangular trust regions are defined around the reference path with distinct lateral and longitudinal thresholds; if the prediction lies within a region at an evaluation time, it receives that rater’s high score, otherwise the score decays with lateral/longitudinal deviations. Furthermore, we add two consistency metrics. Object$\leftrightarrow$Explanation Consistency uses an LLM judge to compare entities in detection versus those referenced in the explanation, reporting two directional errors—\textbf{O} $\rightarrow$ \textbf{E} (detected but not mentioned) and \textbf{E} $\rightarrow$ \textbf{O} (mentioned but not detected)—with higher values indicating better agreement. Behavior$\leftrightarrow$Trajectory Consistency converts the predicted trajectory to a meta action via rules and measures its agreement with the predicted behavior; to discourage trivial ``keep lane/keep speed,'' we compute a weighted accuracy using ADE weights. \looseness=-1

\begin{figure*}[!t]
    \centering
    \includegraphics[width=1.0\linewidth]{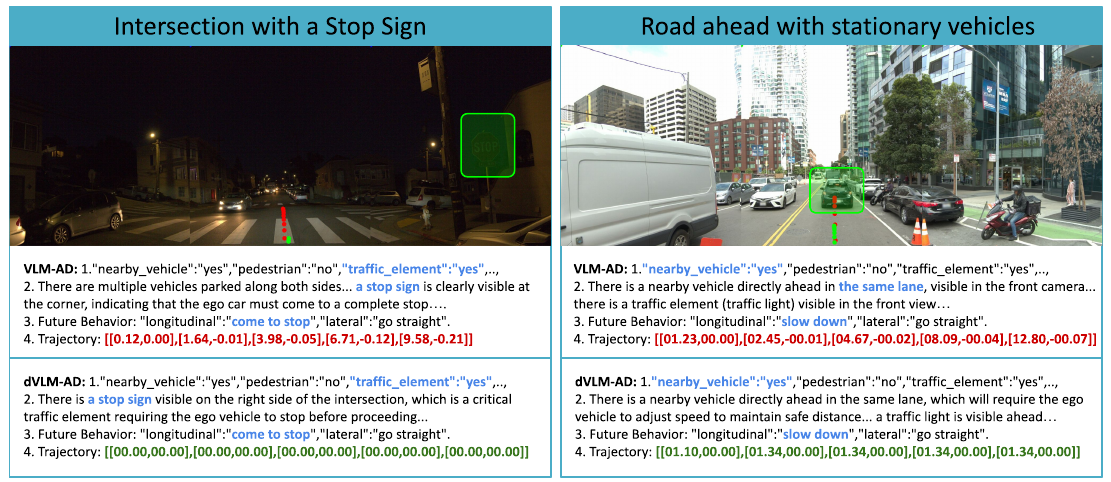}
      \vspace{-3mm}
    \caption{Examples of our \model{} demonstrate stronger consistency between reasoning and action than autoregressive VLMs.}
    \label{fig:demo}
    \vspace{-6mm}
\end{figure*}

\subsection{Main Results} \label{sec:main}

\paragraph{Consistency Analysis.}
As shown in Table~\ref{tab:eval_consistency}, \model{} achieves superior overall consistency compared to autoregressive baselines, highlighting the inherent advantages of diffusion-based reasoning in driving tasks. 
Across both datasets, \textit{Object$\leftrightarrow$Explanation} alignment remains high, with \model{} approaching 99\% accuracy and improving upon the strongest AR baseline by roughly 5\%. 
This indicates that the diffusion model generates explanations that are more faithful to visual evidence and perceived traffic elements. 
For \textit{Behavior$\leftrightarrow$Trajectory}, there is a noticeable asymmetry between longitudinal and lateral alignment. 
Longitudinal reasoning proves more challenging due to AR-based VLMs’ tendency to produce inconsistent velocity plans—such as declaring a \emph{stop} or \emph{slow down} while actually accelerating (Fig.~\ref{fig:demo})—whereas lateral behaviors, typically governed by navigation commands, remain more stable. 
Moreover, nuScenes yields higher overall consistency than WOD-E2E, as its simpler traffic conditions often involve straightforward maneuvers like \emph{go straight} or \emph{keep speed}. 
Finally, under identical model capacity and training data, the diffusion-based \model{} surpasses its autoregressive counterpart (VLM-AD) on planning metrics, reducing collision rate by about 9\% and increasing human-rated RFS by 6\%. 
These results collectively validate that diffusion VLMs serve as a more reliable backbone for achieving semantically consistent and safe autonomous driving plans.

\vspace{-5mm}
\paragraph{WOD-E2E Results.}

Table~\ref{tab:waymo} presents performance comparisons on the WOD-E2E test set, which specifically targets long-tail driving scenarios. Our \model{} obtains the highest RFS (7.633) despite using fewer trajectories (29.3k) and a weaker backbone (LLaDA-V) than the leading AR-based method AutoVLA (RFS 7.557) with 52.8k trajectories and backbone Qwen2.5-VL. Note that the zero-shot models OpenEMMA and LightEMMA achieve substantially lower RFS (5.158 and 6.517 respectively) and much higher ADE@5s, underscoring the difficulty of the benchmark for methods without fine-tuned planning. The superior RFS and lowest ADE@3s achieved by our model indicate that diffusion-based VLMs are not only more human-aligned but also more accurate in short-horizon trajectory forecasting under rare and complex driving conditions.


\vspace{-5mm}
\paragraph{nuScenes Results.}
We evaluate \model{} on the nuScenes validation set, as shown in Table~\ref{tab:nuscenes}. For a fair comparison, baselines are restricted to traditional training-based driving policies and VLM/VLA systems that jointly output both reasoning and trajectory. Under this protocol, \model{} achieves trajectory accuracy comparable to the strongest VLM/VLA while simultaneously providing coupled reasoning–trajectory outputs. Zero-shot VLM/VLA baselines exhibit noticeably larger errors, underscoring the value of structured planning supervision, whereas traditional policies remain competitive on L2 but do not yield explicit reasoning. \looseness=-1

\subsection{Ablation Studies} \label{sec:ablation}

\paragraph{Denoise Strategies.}
As mentioned in Section~\ref{sec:denoise}, we compare a naive fixed-slot denoising scheme with our dynamic denoise strategy under identical backbones, training data, and prompts; only the decoding policy differs. As shown in Table~\ref{tab:denoise}, the naive strategy yields substantially lower accuracy on both longitudinal (46.30) and lateral (26.50) behaviors, averaging 36.40. A diagnostic of the lateral predictions reveals a pronounced length-matching bias: because the template reserves five masked tokens for the lateral slot, the model disproportionately outputs \texttt{right\_lane\_change}-the longest behavior phrase in our meta behavior set—with \textbf{43\%} of lateral predictions collapsing to this single class. In contrast, our method introduces a dynamic denoise policy with a sentinel \textit{reduce token} and a short warm-up, allowing spans to terminate early once sufficient evidence is formed rather than filling all reserved blanks.

\begin{table}[!t]
\centering
\caption{Comparison of different denoise strategies.}
\label{tab:denoise}
\vspace{-3mm}
\small
\resizebox{0.46\textwidth}{!}{
\begin{tabular}{lccc}
\toprule
\textbf{Method} & \textbf{longitudinal} & \textbf{lateral} & \textbf{Avg.} \\
\midrule
Naive & 46.30 & 26.50 & 36.40 \\
Ours  & 74.40~{\textcolor{green!60!black}{(+28.10)}} & 85.70~{\textcolor{green!60!black}{(+59.20)}} & 80.05~{\textcolor{green!60!black}{(+43.65)}} \\
\bottomrule
\end{tabular}}
\vspace{-5mm}
\end{table}

\vspace{-4mm}
\paragraph{Prompt Perturbation.} \label{sec:attack}
We inject harmful prompts directly into the user navigation command, considering two perturbations, \emph{Order Perturbation (OP)} and \emph{Trajectory Omission (TO)}, that explicitly target the decoding pipeline. OP disrupts the mandated step order (e.g., forcing trajectory generation before perception/objects), while TO coerces the model to continue emitting reasoning until the output budget is exhausted, preventing trajectory emission; implementation details are provided in the appendix. As shown in Table~\ref{tab:attack_results}, AR-based \textsc{VLM-AD} exhibits pronounced degradation under both perturbations, whereas the diffusion, template-anchored \model{} maintains planning quality and consistency under any perturbation. These results indicate that schedule- and budget-level prompt manipulations can severely compromise autoregressive planners, while schema-constrained infilling confers robustness in our setting.

\begin{table}[!t]
\caption{Comparison between VLM-AD and \model{} under different prompt perturbations. 
\textit{OP} corresponds to \textbf{Order Perturbation}, 
and \textit{TO} corresponds to \textbf{Trajectory Omission}. 
Values in \textcolor{red!60!black}{(red)} denote the relative degradation compared to the vanilla setting.}
\vspace{-2mm}
\centering
\small
\resizebox{0.48\textwidth}{!}{
\begin{tabular}{lccccc}
\toprule
\textbf{Model} & \textbf{Strategy} & \textbf{RFS}$\uparrow$ & \textbf{ADE~(5s)}$\downarrow$ & \textbf{Cons.}$\uparrow$ & \textbf{SR (\%)}$\downarrow$ \\
\midrule
\multirow{3}{*}{VLM-AD} 
 & Vanilla        & 7.42 & 2.34 & 71.50 & -- \\
 & OP   & 7.16~{\textcolor{red!60!black}{(-0.26)}} & 3.17~{\textcolor{red!60!black}{(+0.83)}} & 68.55~{\textcolor{red!60!black}{(-2.95)}} & 100.0\% \\
 & TO  & 5.39~{\textcolor{red!60!black}{(-2.03)}} & 11.41~{\textcolor{red!60!black}{(+9.07)}} & 40.65~{\textcolor{red!60!black}{(-30.85)}} & 94.89\% \\
\midrule
\multirow{3}{*}{dVLM-AD} 
 & Vanilla        & 7.66 & 2.51 & 80.05 & -- \\
 & OP   & 7.58~{\textcolor{red!60!black}{(-0.08)}} & 2.60~{\textcolor{red!60!black}{(+0.09)}} & 79.80~{\textcolor{red!60!black}{(-0.25)}} & 0.00\% \\
 & TO  & 7.49~{\textcolor{red!60!black}{(-0.17)}} & 2.59~{\textcolor{red!60!black}{(+0.08)}} & 76.55~{\textcolor{red!60!black}{(-3.50)}} & 0.00\% \\
\bottomrule
\end{tabular}}
\vspace{-6mm}
\label{tab:attack_results}
\end{table}

\section{Conclusion}

We present \model{}, a diffusion-based vision–language model addressing two central challenges in driving VLMs: (i) \emph{reasoning–action inconsistency} and (ii) \emph{uncontrollable generation}. By anchoring generation to a schema and refining bidirectionally, \model{} tightly couples the reasoning trace with predicted actions while constraining decoding to the prescribed structure. Empirically, this delivers more reliable reasoning–action consistency and greater robustness to prompt perturbations on nuScenes and WOD-E2E under matched training settings. Despite employing a relatively weak backbone and textual waypoints, \model{} achieves planning performance comparable to existing driving VLMs/VLAs. Our future work will focus on accelerating runtime of dVLMs and adopting action-centric decoders that embed physics priors to further improve reliability. 




\clearpage
{
    \small
    \bibliographystyle{ieeenat_fullname}
    \bibliography{main}

@misc{brohan2023rt2visionlanguageactionmodelstransfer,
      title={RT-2: Vision-Language-Action Models Transfer Web Knowledge to Robotic Control}, 
      author={Anthony Brohan and Noah Brown and Justice Carbajal and Yevgen Chebotar and Xi Chen and Krzysztof Choromanski and Tianli Ding and Danny Driess and Avinava Dubey and Chelsea Finn and Pete Florence and Chuyuan Fu and Montse Gonzalez Arenas and Keerthana Gopalakrishnan and Kehang Han and Karol Hausman and Alexander Herzog and Jasmine Hsu and Brian Ichter and Alex Irpan and Nikhil Joshi and Ryan Julian and Dmitry Kalashnikov and Yuheng Kuang and Isabel Leal and Lisa Lee and Tsang-Wei Edward Lee and Sergey Levine and Yao Lu and Henryk Michalewski and Igor Mordatch and Karl Pertsch and Kanishka Rao and Krista Reymann and Michael Ryoo and Grecia Salazar and Pannag Sanketi and Pierre Sermanet and Jaspiar Singh and Anikait Singh and Radu Soricut and Huong Tran and Vincent Vanhoucke and Quan Vuong and Ayzaan Wahid and Stefan Welker and Paul Wohlhart and Jialin Wu and Fei Xia and Ted Xiao and Peng Xu and Sichun Xu and Tianhe Yu and Brianna Zitkovich},
      year={2023},
      eprint={2307.15818},
      archivePrefix={arXiv},
      primaryClass={cs.RO},
      url={https://arxiv.org/abs/2307.15818}, 
}

@misc{kim2024openvlaopensourcevisionlanguageactionmodel,
      title={OpenVLA: An Open-Source Vision-Language-Action Model}, 
      author={Moo Jin Kim and Karl Pertsch and Siddharth Karamcheti and Ted Xiao and Ashwin Balakrishna and Suraj Nair and Rafael Rafailov and Ethan Foster and Grace Lam and Pannag Sanketi and Quan Vuong and Thomas Kollar and Benjamin Burchfiel and Russ Tedrake and Dorsa Sadigh and Sergey Levine and Percy Liang and Chelsea Finn},
      year={2024},
      eprint={2406.09246},
      archivePrefix={arXiv},
      primaryClass={cs.RO},
      url={https://arxiv.org/abs/2406.09246}, 
}

@misc{intelligence2025pi05visionlanguageactionmodelopenworld,
      title={$\pi_{0.5}$: a Vision-Language-Action Model with Open-World Generalization}, 
      author={Physical Intelligence and Kevin Black and Noah Brown and James Darpinian and Karan Dhabalia and Danny Driess and Adnan Esmail and Michael Equi and Chelsea Finn and Niccolo Fusai and Manuel Y. Galliker and Dibya Ghosh and Lachy Groom and Karol Hausman and Brian Ichter and Szymon Jakubczak and Tim Jones and Liyiming Ke and Devin LeBlanc and Sergey Levine and Adrian Li-Bell and Mohith Mothukuri and Suraj Nair and Karl Pertsch and Allen Z. Ren and Lucy Xiaoyang Shi and Laura Smith and Jost Tobias Springenberg and Kyle Stachowicz and James Tanner and Quan Vuong and Homer Walke and Anna Walling and Haohuan Wang and Lili Yu and Ury Zhilinsky},
      year={2025},
      eprint={2504.16054},
      archivePrefix={arXiv},
      primaryClass={cs.LG},
      url={https://arxiv.org/abs/2504.16054}, 
}

@misc{black2024pi0visionlanguageactionflowmodel,
      title={$\pi_0$: A Vision-Language-Action Flow Model for General Robot Control}, 
      author={Kevin Black and Noah Brown and Danny Driess and Adnan Esmail and Michael Equi and Chelsea Finn and Niccolo Fusai and Lachy Groom and Karol Hausman and Brian Ichter and Szymon Jakubczak and Tim Jones and Liyiming Ke and Sergey Levine and Adrian Li-Bell and Mohith Mothukuri and Suraj Nair and Karl Pertsch and Lucy Xiaoyang Shi and James Tanner and Quan Vuong and Anna Walling and Haohuan Wang and Ury Zhilinsky},
      year={2024},
      eprint={2410.24164},
      archivePrefix={arXiv},
      primaryClass={cs.LG},
      url={https://arxiv.org/abs/2410.24164}, 
}

@misc{yang2025lohovlaunifiedvisionlanguageactionmodel,
      title={LoHoVLA: A Unified Vision-Language-Action Model for Long-Horizon Embodied Tasks}, 
      author={Yi Yang and Jiaxuan Sun and Siqi Kou and Yihan Wang and Zhijie Deng},
      year={2025},
      eprint={2506.00411},
      archivePrefix={arXiv},
      primaryClass={cs.RO},
      url={https://arxiv.org/abs/2506.00411}, 
}

@misc{octomodelteam2024octoopensourcegeneralistrobot,
      title={Octo: An Open-Source Generalist Robot Policy}, 
      author={Octo Model Team and Dibya Ghosh and Homer Walke and Karl Pertsch and Kevin Black and Oier Mees and Sudeep Dasari and Joey Hejna and Tobias Kreiman and Charles Xu and Jianlan Luo and You Liang Tan and Lawrence Yunliang Chen and Pannag Sanketi and Quan Vuong and Ted Xiao and Dorsa Sadigh and Chelsea Finn and Sergey Levine},
      year={2024},
      eprint={2405.12213},
      archivePrefix={arXiv},
      primaryClass={cs.RO},
      url={https://arxiv.org/abs/2405.12213}, 
}

@misc{zhou2025chatvla2visionlanguageactionmodelopenworld,
      title={ChatVLA-2: Vision-Language-Action Model with Open-World Embodied Reasoning from Pretrained Knowledge}, 
      author={Zhongyi Zhou and Yichen Zhu and Junjie Wen and Chaomin Shen and Yi Xu},
      year={2025},
      eprint={2505.21906},
      archivePrefix={arXiv},
      primaryClass={cs.RO},
      url={https://arxiv.org/abs/2505.21906}, 
}

@misc{li2025cogvlacognitionalignedvisionlanguageactionmodel,
      title={CogVLA: Cognition-Aligned Vision-Language-Action Model via Instruction-Driven Routing \& Sparsification}, 
      author={Wei Li and Renshan Zhang and Rui Shao and Jie He and Liqiang Nie},
      year={2025},
      eprint={2508.21046},
      archivePrefix={arXiv},
      primaryClass={cs.CV},
      url={https://arxiv.org/abs/2508.21046}, 
}

@misc{hung2025norasmallopensourcedgeneralist,
      title={NORA: A Small Open-Sourced Generalist Vision Language Action Model for Embodied Tasks}, 
      author={Chia-Yu Hung and Qi Sun and Pengfei Hong and Amir Zadeh and Chuan Li and U-Xuan Tan and Navonil Majumder and Soujanya Poria},
      year={2025},
      eprint={2504.19854},
      archivePrefix={arXiv},
      primaryClass={cs.RO},
      url={https://arxiv.org/abs/2504.19854}, 
}

@misc{xing2025openemmaopensourcemultimodalmodel,
      title={OpenEMMA: Open-Source Multimodal Model for End-to-End Autonomous Driving}, 
      author={Shuo Xing and Chengyuan Qian and Yuping Wang and Hongyuan Hua and Kexin Tian and Yang Zhou and Zhengzhong Tu},
      year={2025},
      eprint={2412.15208},
      archivePrefix={arXiv},
      primaryClass={cs.CV},
      url={https://arxiv.org/abs/2412.15208}, 
}

@misc{zhou2025opendrivevlaendtoendautonomousdriving,
      title={OpenDriveVLA: Towards End-to-end Autonomous Driving with Large Vision Language Action Model}, 
      author={Xingcheng Zhou and Xuyuan Han and Feng Yang and Yunpu Ma and Alois C. Knoll},
      year={2025},
      eprint={2503.23463},
      archivePrefix={arXiv},
      primaryClass={cs.CV},
      url={https://arxiv.org/abs/2503.23463}, 
}

@misc{chi2025impromptuvlaopenweights,
      title={Impromptu VLA: Open Weights and Open Data for Driving Vision-Language-Action Models}, 
      author={Haohan Chi and Huan-ang Gao and Ziming Liu and Jianing Liu and Chenyu Liu and Jinwei Li and Kaisen Yang and Yangcheng Yu and Zeda Wang and Wenyi Li and Leichen Wang and Xingtao Hu and Hao Sun and Hang Zhao and Hao Zhao},
      year={2025},
      eprint={2505.23757},
      archivePrefix={arXiv},
      primaryClass={cs.CV},
      url={https://arxiv.org/abs/2505.23757}, 
}

@misc{zhou2025autovlavisionlanguageactionmodelendtoend,
      title={AutoVLA: A Vision-Language-Action Model for End-to-End Autonomous Driving with Adaptive Reasoning and Reinforcement Fine-Tuning}, 
      author={Zewei Zhou and Tianhui Cai and Seth Z. Zhao and Yun Zhang and Zhiyu Huang and Bolei Zhou and Jiaqi Ma},
      year={2025},
      eprint={2506.13757},
      archivePrefix={arXiv},
      primaryClass={cs.CV},
      url={https://arxiv.org/abs/2506.13757}, 
}

@article{jiang2025diffvla,
  title={Diffvla: Vision-language guided diffusion planning for autonomous driving},
  author={Jiang, Anqing and Gao, Yu and Sun, Zhigang and Wang, Yiru and Wang, Jijun and Chai, Jinghao and Cao, Qian and Heng, Yuweng and Jiang, Hao and Dong, Yunda and others},
  journal={arXiv preprint arXiv:2505.19381},
  year={2025}
}

@article{tian2024tokenize,
  title={Tokenize the world into object-level knowledge to address long-tail events in autonomous driving},
  author={Tian, Ran and Li, Boyi and Weng, Xinshuo and Chen, Yuxiao and Schmerling, Edward and Wang, Yue and Ivanovic, Boris and Pavone, Marco},
  journal={arXiv preprint arXiv:2407.00959},
  year={2024}
}

@inproceedings{li2024ego,
  title={Is ego status all you need for open-loop end-to-end autonomous driving?},
  author={Li, Zhiqi and Yu, Zhiding and Lan, Shiyi and Li, Jiahan and Kautz, Jan and Lu, Tong and Alvarez, Jose M},
  booktitle={Proceedings of the IEEE/CVF Conference on Computer Vision and Pattern Recognition},
  pages={14864--14873},
  year={2024}
}

@inproceedings{jiang2023vad,
  title={Vad: Vectorized scene representation for efficient autonomous driving},
  author={Jiang, Bo and Chen, Shaoyu and Xu, Qing and Liao, Bencheng and Chen, Jiajie and Zhou, Helong and Zhang, Qian and Liu, Wenyu and Huang, Chang and Wang, Xinggang},
  booktitle={Proceedings of the IEEE/CVF International Conference on Computer Vision},
  pages={8340--8350},
  year={2023}
}

@inproceedings{hu2023planning,
  title={Planning-oriented autonomous driving},
  author={Hu, Yihan and Yang, Jiazhi and Chen, Li and Li, Keyu and Sima, Chonghao and Zhu, Xizhou and Chai, Siqi and Du, Senyao and Lin, Tianwei and Wang, Wenhai and others},
  booktitle={Proceedings of the IEEE/CVF conference on computer vision and pattern recognition},
  pages={17853--17862},
  year={2023}
}

@article{tian2024drivevlm,
  title={Drivevlm: The convergence of autonomous driving and large vision-language models},
  author={Tian, Xiaoyu and Gu, Junru and Li, Bailin and Liu, Yicheng and Wang, Yang and Zhao, Zhiyong and Zhan, Kun and Jia, Peng and Lang, Xianpeng and Zhao, Hang},
  journal={arXiv preprint arXiv:2402.12289},
  year={2024}
}

@article{hwang2024emma,
  title={Emma: End-to-end multimodal model for autonomous driving},
  author={Hwang, Jyh-Jing and Xu, Runsheng and Lin, Hubert and Hung, Wei-Chih and Ji, Jingwei and Choi, Kristy and Huang, Di and He, Tong and Covington, Paul and Sapp, Benjamin and others},
  journal={arXiv preprint arXiv:2410.23262},
  year={2024}
}

@article{zhou2025opendrivevla,
  title={Opendrivevla: Towards end-to-end autonomous driving with large vision language action model},
  author={Zhou, Xingcheng and Han, Xuyuan and Yang, Feng and Ma, Yunpu and Knoll, Alois C},
  journal={arXiv preprint arXiv:2503.23463},
  year={2025}
}

@article{alayrac2022flamingo,
  title={Flamingo: a visual language model for few-shot learning},
  author={Alayrac, Jean-Baptiste and Donahue, Jeff and Luc, Pauline and Miech, Antoine and Barr, Iain and Hasson, Yana and Lenc, Karel and Mensch, Arthur and Millican, Katherine and Reynolds, Malcolm and others},
  journal={Advances in neural information processing systems},
  volume={35},
  pages={23716--23736},
  year={2022}
}

@inproceedings{liu2024improved,
  title={Improved baselines with visual instruction tuning},
  author={Liu, Haotian and Li, Chunyuan and Li, Yuheng and Lee, Yong Jae},
  booktitle={Proceedings of the IEEE/CVF conference on computer vision and pattern recognition},
  pages={26296--26306},
  year={2024}
}

@misc{nvidia2025alpamayor1bridgingreasoningaction,
      title={Alpamayo-R1: Bridging Reasoning and Action Prediction for Generalizable Autonomous Driving in the Long Tail}, 
      author={NVIDIA and : and Yan Wang and Wenjie Luo and Junjie Bai and Yulong Cao and Tong Che and Ke Chen and Yuxiao Chen and Jenna Diamond and Yifan Ding and Wenhao Ding and Liang Feng and Greg Heinrich and Jack Huang and Peter Karkus and Boyi Li and Pinyi Li and Tsung-Yi Lin and Dongran Liu and Ming-Yu Liu and Langechuan Liu and Zhijian Liu and Jason Lu and Yunxiang Mao and Pavlo Molchanov and Lindsey Pavao and Zhenghao Peng and Mike Ranzinger and Ed Schmerling and Shida Shen and Yunfei Shi and Sarah Tariq and Ran Tian and Tilman Wekel and Xinshuo Weng and Tianjun Xiao and Eric Yang and Xiaodong Yang and Yurong You and Xiaohui Zeng and Wenyuan Zhang and Boris Ivanovic and Marco Pavone},
      year={2025},
      eprint={2511.00088},
      archivePrefix={arXiv},
      primaryClass={cs.RO},
      url={https://arxiv.org/abs/2511.00088}, 
}

@online{waymo_e2e_2025,
  author   = {{Waymo LLC}},
  title    = {Waymo Open Dataset: 2025 End-to-End Driving Challenge},
  year     = {2025},
  url      = {https://waymo.com/open/challenges/2025/e2e-driving/},
  urldate  = {2025-11-12},
  note     = {Accessed: November 12, 2025}
}

@article{qiao2025lightemma,
  title={Lightemma: Lightweight end-to-end multimodal model for autonomous driving},
  author={Qiao, Zhijie and Li, Haowei and Cao, Zhong and Liu, Henry X},
  journal={arXiv preprint arXiv:2505.00284},
  year={2025}
}

@online{openai_gpt41_2025,
  author   = {OpenAI},
  title    = {Introducing GPT-4.1 in the API},
  year     = {2025},
  month    = apr,
  url      = {https://openai.com/index/gpt-4-1/},
  urldate  = {2025-11-12},
  note     = {OpenAI}
}

@misc{qwen3technicalreport,
      title={Qwen3 Technical Report}, 
      author={Qwen Team},
      year={2025},
      eprint={2505.09388},
      archivePrefix={arXiv},
      primaryClass={cs.CL},
      url={https://arxiv.org/abs/2505.09388}, 
}

@article{wang2025internvl3,
  title={Internvl3. 5: Advancing open-source multimodal models in versatility, reasoning, and efficiency},
  author={Wang, Weiyun and Gao, Zhangwei and Gu, Lixin and Pu, Hengjun and Cui, Long and Wei, Xingguang and Liu, Zhaoyang and Jing, Linglin and Ye, Shenglong and Shao, Jie and others},
  journal={arXiv preprint arXiv:2508.18265},
  year={2025}
}

@article{kuo2025h,
  title={H-cot: Hijacking the chain-of-thought safety reasoning mechanism to jailbreak large reasoning models, including openai o1/o3, deepseek-r1, and gemini 2.0 flash thinking},
  author={Kuo, Martin and Zhang, Jianyi and Ding, Aolin and Wang, Qinsi and DiValentin, Louis and Bao, Yujia and Wei, Wei and Li, Hai and Chen, Yiran},
  journal={arXiv preprint arXiv:2502.12893},
  year={2025}
}

@article{you2025llada,
  title={Llada-v: Large language diffusion models with visual instruction tuning},
  author={You, Zebin and Nie, Shen and Zhang, Xiaolu and Hu, Jun and Zhou, Jun and Lu, Zhiwu and Wen, Ji-Rong and Li, Chongxuan},
  journal={arXiv preprint arXiv:2505.16933},
  year={2025}
}

@article{yu2025dimple,
  title={Dimple: Discrete diffusion multimodal large language model with parallel decoding},
  author={Yu, Runpeng and Ma, Xinyin and Wang, Xinchao},
  journal={arXiv preprint arXiv:2505.16990},
  year={2025}
}

@article{gulrajani2023likelihood,
  title={Likelihood-based diffusion language models},
  author={Gulrajani, Ishaan and Hashimoto, Tatsunori B},
  journal={Advances in Neural Information Processing Systems},
  volume={36},
  pages={16693--16715},
  year={2023}
}

@article{hao2025driveaction,
  title={Driveaction: A benchmark for exploring human-like driving decisions in vla models},
  author={Hao, Yuhan and Li, Zhengning and Sun, Lei and Wang, Weilong and Yi, Naixin and Song, Sheng and Qin, Caihong and Zhou, Mofan and Zhan, Yifei and Lang, Xianpeng},
  journal={arXiv preprint arXiv:2506.05667},
  year={2025}
}

@article{xu2025wod,
  title={WOD-E2E: Waymo Open Dataset for End-to-End Driving in Challenging Long-tail Scenarios},
  author={Xu, Runsheng and Lin, Hubert and Jeon, Wonseok and Feng, Hao and Zou, Yuliang and Sun, Liting and Gorman, John and Tolstaya, Kate and Tang, Sarah and White, Brandyn and others},
  journal={arXiv preprint arXiv:2510.26125},
  year={2025}
}

@article{luo2025adathinkdrive,
  title={AdaThinkDrive: Adaptive Thinking via Reinforcement Learning for Autonomous Driving},
  author={Luo, Yuechen and Li, Fang and Xu, Shaoqing and Lai, Zhiyi and Yang, Lei and Chen, Qimao and Luo, Ziang and Xie, Zixun and Jiang, Shengyin and Liu, Jiaxin and others},
  journal={arXiv preprint arXiv:2509.13769},
  year={2025}
}

@article{bai2025qwen2,
  title={Qwen2. 5-vl technical report},
  author={Bai, Shuai and Chen, Keqin and Liu, Xuejing and Wang, Jialin and Ge, Wenbin and Song, Sibo and Dang, Kai and Wang, Peng and Wang, Shijie and Tang, Jun and others},
  journal={arXiv preprint arXiv:2502.13923},
  year={2025}
}

@article{wang2024qwen2,
  title={Qwen2-vl: Enhancing vision-language model's perception of the world at any resolution},
  author={Wang, Peng and Bai, Shuai and Tan, Sinan and Wang, Shijie and Fan, Zhihao and Bai, Jinze and Chen, Keqin and Liu, Xuejing and Wang, Jialin and Ge, Wenbin and others},
  journal={arXiv preprint arXiv:2409.12191},
  year={2024}
}

@article{tschannen2025siglip,
  title={Siglip 2: Multilingual vision-language encoders with improved semantic understanding, localization, and dense features},
  author={Tschannen, Michael and Gritsenko, Alexey and Wang, Xiao and Naeem, Muhammad Ferjad and Alabdulmohsin, Ibrahim and Parthasarathy, Nikhil and Evans, Talfan and Beyer, Lucas and Xia, Ye and Mustafa, Basil and others},
  journal={arXiv preprint arXiv:2502.14786},
  year={2025}
}

@article{nie2025large,
  title={Large language diffusion models},
  author={Nie, Shen and Zhu, Fengqi and You, Zebin and Zhang, Xiaolu and Ou, Jingyang and Hu, Jun and Zhou, Jun and Lin, Yankai and Wen, Ji-Rong and Li, Chongxuan},
  journal={arXiv preprint arXiv:2502.09992},
  year={2025}
}

@article{yuan2025autodrive,
  title={Autodrive-R$^2$: Incentivizing reasoning and self-reflection capacity for VLA model in autonomous driving},
  author={Yuan, Zhenlong and Tang, Jing and Luo, Jinguo and Chen, Rui and Qian, Chengxuan and Sun, Lei and Chu, Xiangxiang and Cai, Yujun and Zhang, Dapeng and Li, Shuo},
  journal={arXiv preprint arXiv:2509.01944},
  year={2025}
}

@article{xiong2025unveiling,
  title={Unveiling the Potential of Diffusion Large Language Model in Controllable Generation},
  author={Xiong, Zhen and Cai, Yujun and Li, Zhecheng and Wang, Yiwei},
  journal={arXiv preprint arXiv:2507.04504},
  year={2025}
}

@misc{ishaq2025drivelmmo1stepbystepreasoningdataset,
      title={DriveLMM-o1: A Step-by-Step Reasoning Dataset and Large Multimodal Model for Driving Scenario Understanding}, 
      author={Ayesha Ishaq and Jean Lahoud and Ketan More and Omkar Thawakar and Ritesh Thawkar and Dinura Dissanayake and Noor Ahsan and Yuhao Li and Fahad Shahbaz Khan and Hisham Cholakkal and Ivan Laptev and Rao Muhammad Anwer and Salman Khan},
      year={2025},
      eprint={2503.10621},
      archivePrefix={arXiv},
      primaryClass={cs.CV},
      url={https://arxiv.org/abs/2503.10621}, 
}

@article{li2025diffusion,
  title={Diffusion language models know the answer before decoding},
  author={Li, Pengxiang and Zhou, Yefan and Muhtar, Dilxat and Yin, Lu and Yan, Shilin and Shen, Li and Liang, Yi and Vosoughi, Soroush and Liu, Shiwei},
  journal={arXiv preprint arXiv:2508.19982},
  year={2025}
}

@article{li2025beyond,
  title={Beyond fixed: Training-free variable-length denoising for diffusion large language models},
  author={Li, Jinsong and Dong, Xiaoyi and Zang, Yuhang and Cao, Yuhang and Wang, Jiaqi and Lin, Dahua},
  journal={arXiv preprint arXiv:2508.00819},
  year={2025}
}

@article{zeng2025futuresightdrive,
  title={FutureSightDrive: Thinking Visually with Spatio-Temporal CoT for Autonomous Driving},
  author={Zeng, Shuang and Chang, Xinyuan and Xie, Mengwei and Liu, Xinran and Bai, Yifan and Pan, Zheng and Xu, Mu and Wei, Xing},
  journal={arXiv preprint arXiv:2505.17685},
  year={2025}
}

@article{li2025recogdrive,
  title={Recogdrive: A reinforced cognitive framework for end-to-end autonomous driving},
  author={Li, Yongkang and Xiong, Kaixin and Guo, Xiangyu and Li, Fang and Yan, Sixu and Xu, Gangwei and Zhou, Lijun and Chen, Long and Sun, Haiyang and Wang, Bing and others},
  journal={arXiv preprint arXiv:2506.08052},
  year={2025}
}

@article{rowe2025poutine,
  title={Poutine: Vision-Language-Trajectory Pre-Training and Reinforcement Learning Post-Training Enable Robust End-to-End Autonomous Driving},
  author={Rowe, Luke and de Schaetzen, Rodrigue and Girgis, Roger and Pal, Christopher and Paull, Liam},
  journal={arXiv preprint arXiv:2506.11234},
  year={2025}
}

@inproceedings{caesar2020nuscenes,
  title={nuscenes: A multimodal dataset for autonomous driving},
  author={Caesar, Holger and Bankiti, Varun and Lang, Alex H and Vora, Sourabh and Liong, Venice Erin and Xu, Qiang and Krishnan, Anush and Pan, Yu and Baldan, Giancarlo and Beijbom, Oscar},
  booktitle={Proceedings of the IEEE/CVF conference on computer vision and pattern recognition},
  pages={11621--11631},
  year={2020}
}

@inproceedings{sima2024drivelm,
  title={Drivelm: Driving with graph visual question answering},
  author={Sima, Chonghao and Renz, Katrin and Chitta, Kashyap and Chen, Li and Zhang, Hanxue and Xie, Chengen and Bei{\ss}wenger, Jens and Luo, Ping and Geiger, Andreas and Li, Hongyang},
  booktitle={European conference on computer vision},
  pages={256--274},
  year={2024},
  organization={Springer}
}

@article{wang2024drivecot,
  title={Drivecot: Integrating chain-of-thought reasoning with end-to-end driving},
  author={Wang, Tianqi and Xie, Enze and Chu, Ruihang and Li, Zhenguo and Luo, Ping},
  journal={arXiv preprint arXiv:2403.16996},
  year={2024}
}

@article{li2022diffusion,
  title={Diffusion-lm improves controllable text generation},
  author={Li, Xiang and Thickstun, John and Gulrajani, Ishaan and Liang, Percy S and Hashimoto, Tatsunori B},
  journal={Advances in neural information processing systems},
  volume={35},
  pages={4328--4343},
  year={2022}
}

@article{liang2025discrete,
  title={Discrete diffusion vla: Bringing discrete diffusion to action decoding in vision-language-action policies},
  author={Liang, Zhixuan and Li, Yizhuo and Yang, Tianshuo and Wu, Chengyue and Mao, Sitong and Pei, Liuao and Yang, Xiaokang and Pang, Jiangmiao and Mu, Yao and Luo, Ping},
  journal={arXiv preprint arXiv:2508.20072},
  year={2025}
}

@article{li2025discrete,
  title={Discrete diffusion for reflective vision-language-action models in autonomous driving},
  author={Li, Pengxiang and Zheng, Yinan and Wang, Yue and Wang, Huimin and Zhao, Hang and Liu, Jingjing and Zhan, Xianyuan and Zhan, Kun and Lang, Xianpeng},
  journal={arXiv preprint arXiv:2509.20109},
  year={2025}
}

@article{xu2024vlm,
  title={Vlm-ad: End-to-end autonomous driving through vision-language model supervision},
  author={Xu, Yi and Hu, Yuxin and Zhang, Zaiwei and Meyer, Gregory P and Mustikovela, Siva Karthik and Srinivasa, Siddhartha and Wolff, Eric M and Huang, Xin},
  journal={arXiv preprint arXiv:2412.14446},
  year={2024}
}

@article{wen2025dvla,
  title={dVLA: Diffusion Vision-Language-Action Model with Multimodal Chain-of-Thought},
  author={Wen, Junjie and Zhu, Minjie and Liu, Jiaming and Liu, Zhiyuan and Yang, Yicun and Zhang, Linfeng and Zhang, Shanghang and Zhu, Yichen and Xu, Yi},
  journal={arXiv preprint arXiv:2509.25681},
  year={2025}
}

@article{wen2025llada,
  title={Llada-vla: Vision language diffusion action models},
  author={Wen, Yuqing and Li, Hebei and Gu, Kefan and Zhao, Yucheng and Wang, Tiancai and Sun, Xiaoyan},
  journal={arXiv preprint arXiv:2509.06932},
  year={2025}
}

@article{huang2025survey,
  title={A survey on hallucination in large language models: Principles, taxonomy, challenges, and open questions},
  author={Huang, Lei and Yu, Weijiang and Ma, Weitao and Zhong, Weihong and Feng, Zhangyin and Wang, Haotian and Chen, Qianglong and Peng, Weihua and Feng, Xiaocheng and Qin, Bing and others},
  journal={ACM Transactions on Information Systems},
  volume={43},
  number={2},
  pages={1--55},
  year={2025},
  publisher={ACM New York, NY}
}

@misc{deepmind2025geminidiffusion,
  title        = {Gemini Diffusion: Our state-of-the-art, experimental text diffusion model},
  author       = {{DeepMind}},
  year         = {2025},
  howpublished = {\url{https://deepmind.google/models/gemini-diffusion/}},
  note         = {Accessed: 2025-11-06}
}

@article{yang2025mmada,
  title={Mmada: Multimodal large diffusion language models},
  author={Yang, Ling and Tian, Ye and Li, Bowen and Zhang, Xinchen and Shen, Ke and Tong, Yunhai and Wang, Mengdi},
  journal={arXiv preprint arXiv:2505.15809},
  year={2025}
}

@article{ye2025dream,
  title={Dream 7b: Diffusion large language models},
  author={Ye, Jiacheng and Xie, Zhihui and Zheng, Lin and Gao, Jiahui and Wu, Zirui and Jiang, Xin and Li, Zhenguo and Kong, Lingpeng},
  journal={arXiv preprint arXiv:2508.15487},
  year={2025}
}

@article{guo2025deepseek,
  title={Deepseek-r1: Incentivizing reasoning capability in llms via reinforcement learning},
  author={Guo, Daya and Yang, Dejian and Zhang, Haowei and Song, Junxiao and Zhang, Ruoyu and Xu, Runxin and Zhu, Qihao and Ma, Shirong and Wang, Peiyi and Bi, Xiao and others},
  journal={arXiv preprint arXiv:2501.12948},
  year={2025}
}

@inproceedings{robey2025jailbreaking,
  title={Jailbreaking llm-controlled robots},
  author={Robey, Alexander and Ravichandran, Zachary and Kumar, Vijay and Hassani, Hamed and Pappas, George J},
  booktitle={2025 IEEE International Conference on Robotics and Automation (ICRA)},
  pages={11948--11956},
  year={2025},
  organization={IEEE}
}

@misc{huang2025pathdriftlargereasoning,
      title={Path Drift in Large Reasoning Models:How First-Person Commitments Override Safety}, 
      author={Yuyi Huang and Runzhe Zhan and Lidia S. Chao and Ailin Tao and Derek F. Wong},
      year={2025},
      eprint={2510.10013},
      archivePrefix={arXiv},
      primaryClass={cs.CL},
      url={https://arxiv.org/abs/2510.10013}, 
}

@article{liu2023visual,
  title={Visual instruction tuning},
  author={Liu, Haotian and Li, Chunyuan and Wu, Qingyang and Lee, Yong Jae},
  journal={Advances in neural information processing systems},
  volume={36},
  pages={34892--34916},
  year={2023}
}

@article{li2024llava,
  title={Llava-onevision: Easy visual task transfer},
  author={Li, Bo and Zhang, Yuanhan and Guo, Dong and Zhang, Renrui and Li, Feng and Zhang, Hao and Zhang, Kaichen and Zhang, Peiyuan and Li, Yanwei and Liu, Ziwei and others},
  journal={arXiv preprint arXiv:2408.03326},
  year={2024}
}

@article{zhou2025autovla,
  title={AutoVLA: A Vision-Language-Action Model for End-to-End Autonomous Driving with Adaptive Reasoning and Reinforcement Fine-Tuning},
  author={Zhou, Zewei and Cai, Tianhui and Zhao, Seth Z and Zhang, Yun and Huang, Zhiyu and Zhou, Bolei and Ma, Jiaqi},
  journal={arXiv preprint arXiv:2506.13757},
  year={2025}
}

@article{chi2025impromptu,
  title={Impromptu VLA: Open Weights and Open Data for Driving Vision-Language-Action Models},
  author={Chi, Haohan and Gao, Huan-ang and Liu, Ziming and Liu, Jianing and Liu, Chenyu and Li, Jinwei and Yang, Kaisen and Yu, Yangcheng and Wang, Zeda and Li, Wenyi and others},
  journal={arXiv preprint arXiv:2505.23757},
  year={2025}
}

@article{jiang2025irl,
  title={Irl-vla: Training an vision-language-action policy via reward world model},
  author={Jiang, Anqing and Gao, Yu and Wang, Yiru and Sun, Zhigang and Wang, Shuo and Heng, Yuwen and Sun, Hao and Tang, Shichen and Zhu, Lijuan and Chai, Jinhao and others},
  journal={arXiv preprint arXiv:2508.06571},
  year={2025}
}

@inproceedings{arai2025covla,
  title={Covla: Comprehensive vision-language-action dataset for autonomous driving},
  author={Arai, Hidehisa and Miwa, Keita and Sasaki, Kento and Watanabe, Kohei and Yamaguchi, Yu and Aoki, Shunsuke and Yamamoto, Issei},
  booktitle={2025 IEEE/CVF Winter Conference on Applications of Computer Vision (WACV)},
  pages={1933--1943},
  year={2025},
  organization={IEEE}
}

@inproceedings{
austin2021structured,
title={Structured Denoising Diffusion Models in Discrete State-Spaces},
author={Jacob Austin and Daniel D. Johnson and Jonathan Ho and Daniel Tarlow and Rianne van den Berg},
booktitle={Advances in Neural Information Processing Systems},
editor={A. Beygelzimer and Y. Dauphin and P. Liang and J. Wortman Vaughan},
year={2021},
url={https://openreview.net/forum?id=h7-XixPCAL}
}

@inproceedings{
lou2024discrete,
title={Discrete Diffusion Modeling by Estimating the Ratios of the Data Distribution},
author={Aaron Lou and Chenlin Meng and Stefano Ermon},
booktitle={Forty-first International Conference on Machine Learning},
year={2024},
url={https://openreview.net/forum?id=CNicRIVIPA}
}

@inproceedings{
sahoo2024simple,
title={Simple and Effective Masked Diffusion Language Models},
author={Subham Sekhar Sahoo and Marianne Arriola and Aaron Gokaslan and Edgar Mariano Marroquin and Alexander M Rush and Yair Schiff and Justin T Chiu and Volodymyr Kuleshov},
booktitle={The Thirty-eighth Annual Conference on Neural Information Processing Systems},
year={2024},
url={https://openreview.net/forum?id=L4uaAR4ArM}
}

@inproceedings{
shi2024simplified,
title={Simplified and Generalized Masked Diffusion for Discrete Data},
author={Jiaxin Shi and Kehang Han and Zhe Wang and Arnaud Doucet and Michalis Titsias},
booktitle={The Thirty-eighth Annual Conference on Neural Information Processing Systems},
year={2024},
url={https://openreview.net/forum?id=xcqSOfHt4g}
}

@misc{zhu2025llada15variancereducedpreference,
      title={LLaDA 1.5: Variance-Reduced Preference Optimization for Large Language Diffusion Models}, 
      author={Fengqi Zhu and Rongzhen Wang and Shen Nie and Xiaolu Zhang and Chunwei Wu and Jun Hu and Jun Zhou and Jianfei Chen and Yankai Lin and Ji-Rong Wen and Chongxuan Li},
      year={2025},
      eprint={2505.19223},
      archivePrefix={arXiv},
      primaryClass={cs.LG},
      url={https://arxiv.org/abs/2505.19223}, 
}

@misc{wang2025revolutionizingreinforcementlearningframework,
      title={Revolutionizing Reinforcement Learning Framework for Diffusion Large Language Models}, 
      author={Yinjie Wang and Ling Yang and Bowen Li and Ye Tian and Ke Shen and Mengdi Wang},
      year={2025},
      eprint={2509.06949},
      archivePrefix={arXiv},
      primaryClass={cs.CL},
      url={https://arxiv.org/abs/2509.06949}, 
}
}
\end{document}